%% file: main.tex
\documentclass[10pt,twocolumn,letterpaper]{article}

\usepackage[pagenumbers]{wacv} 
\usepackage{graphicx}
\usepackage{amsmath}
\usepackage{amssymb}
\usepackage{booktabs}
\usepackage{tikz}
\usetikzlibrary{arrows.meta, positioning, calc}
\usepackage[pagebackref,breaklinks,colorlinks]{hyperref}
\usepackage[capitalize]{cleveref}
\usepackage{graphicx}
\usepackage{caption}
\usepackage{cuted}

\def\wacvPaperID{\#2} 
\def\confName{WACV}
\def\confYear{2027}

\begin{document}

\title{Direct Regression and Geometric Recovery for \\ Event-based Spacecraft Pose Estimation}


\author{Pruthvi Nandan Janga \quad
\quad Kashyap Hegde Kota \quad Bharatesh Chakravarthi
\\
Arizona State University\\
{\tt\small pjanga@asu.edu, bharatesh@asu.edu}}

\maketitle


\input{sec/0_abstract}
\input{sec/1_intro}
\input{sec/2_related}
\input{sec/3_spades}

\input{sec/4_direct}
\input{sec/5_modular}
\input{sec/6_experiments}
\input{sec/7_conclusion}

{\small
\bibliographystyle{ieeenat_fullname}
\bibliography{main}
}

\end{document}

%% file: sec/0_abstract.tex
\begin{abstract}

Event cameras offer high temporal resolution and dynamic range for uncooperative spacecraft pose estimation, but models trained on synthetic event streams must generalize to the noise, background activity, and contrast variation of physical sensors. We present a comparative systems study of two independently developed approaches to synthetic-to-real, event-based 6-DoF spacecraft pose estimation. The first is a modular pipeline that combines learned keypoint detection and unlabeled target-domain adaptation with geometry-based pose recovery and temporal smoothing. The second directly regresses translation and rotation from global and cropped event representations using an apparent-scale cue, physics-motivated synthetic corruptions, and feature-consistency regularization. We evaluate both systems under the same SPARK 2026 hidden-test protocol on the real-event split of SPADES, with predictions for all test frames. The results reveal a component-wise reversal concealed by the aggregate challenge score: direct regression reduces orientation error by 22.7\% (1.1767 to 0.9101 rad) and obtains a 13.4\% lower composite score, whereas the modular pipeline reduces relative translation error by 78.7\% (0.1335 to 0.0285). Archived configurations and a qualitative failure case identify plausible explanations for these complementary error profiles, without isolating their causes. These findings motivate a testable hybrid design that combines direct orientation estimation with geometry-constrained translation recovery.

\end{abstract}

%% file: sec/1_intro.tex
\section{Introduction}
\label{sec:intro}

Autonomous spacecraft operations, such as in-orbit docking and active space debris removal, require precise spatial awareness. $6$-DoF pose estimation is the foundational step for these critical maneuvers, accurately determining the $3$D translation and orientation of a target. Historically, traditional frame-based cameras have been the standard for space-based computer vision applications.
However, frame-based cameras can be challenged by the large illumination variations and rapid apparent motion that can be encountered in space. Long exposure relative to target motion can produce blur, while glare, shadows and limited sensor dynamic range can degrade conventional features. Event cameras provide a complementary sensing modality where they report per-pixel asynchronous brightness changes with low latency and high dynamic range~\cite{gallego2020event, chakravarthi2024recent} and has been adopted in various domains \cite{verma2026event, ramesh2026enavi, vinod2025sebvs, tan2025real, chakravarthi2023event, chanda2025event, chanda2025sepose, imandi2026adafuse}. These properties can reduce motion blur and extend operation across difficult illumination conditions, although event sensors remain affected by noise, contrast thresholds and background activity. They are therefore more promising than universally preferable sensors for spacecraft pose estimation.

Despite the hardware advantages of event sensors, deploying event-based neural networks introduces a severe sim-to-real domain gap. Due to the high cost and logistical constraints of capturing annotated real-world orbital event data, models are often trained exclusively on synthetic trajectories. When these models are evaluated on real-world event streams, such as the physical laboratory sequences in the SPADES dataset, they sharply decline in performance, primarily due to the domain gap caused by asynchronous sensor noise, dead pixels, and background artifacts. This domain shift is illustrated in Figure~\ref{fig:SPADES}: simulated
events preserve relatively clean spacecraft structure, whereas the
hardware recording contains stronger background activity and sensor
artifacts that can disrupt both apparent-scale cues and keypoint
localization.
The domain gap affects cues used for the two pose components differently. Monocular translation, especially along the optical axis, benefits from global apparent scale cues, whereas rotation estimation benefits from localized structural evidence. Real event noise can perturb both the spatial extent of the event cloud and the edges used for keypoint localization. The two most common system designs address these requirements differently:

\begin{figure*}[t]
  \centering
  \includegraphics[width=1\linewidth]{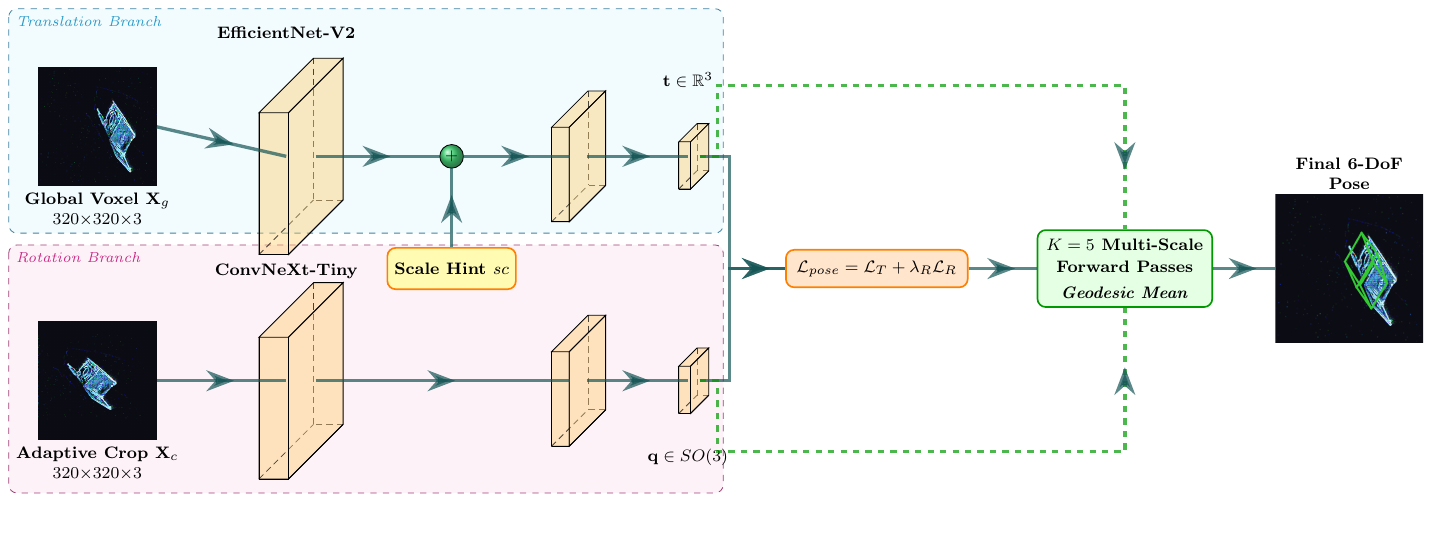}
  \caption{Direct regression system - separate branches process a global voxel for translation, an adaptive crop for rotation, and event-spread feature provides an additional apparent-scale cue.}
  \label{fig:direct-pipeline}
\end{figure*}

\begin{enumerate}
\item \textbf{\textit{Modular keypoint and geometry:}} A feature extractor predicts $2$D spacecraft keypoints, which are paired with known $3$D locations for perspective-n-point (PnP) pose recovery~\cite{chen2019satellite,item_34721c4a90c2421c9df615a96127a4a2}. This decomposition is interpretable and enforces object geometry but corrupted keypoints can confuse the solver.
\item \textbf{\textit{End-to-end direct regression paradigm:}} A neural network predicts translation and rotation from an event representation~\cite{proenca2020deep}. Global features can use contextual evidence without explicit correspondences but monocular translation remains sensitive to how the network represents apparent scale.
\end{enumerate}

This paper presents two independently developed systems for spacecraft pose estimation and relative navigation on the SPADES event-camera benchmark~\cite{rathinam2024spades}. The modular system (see Figure~\ref{fig:modular-pipeline}) combines HRNet keypoint prediction, unlabeled target-domain adaptation, EPnP with RANSAC, and temporal smoothing. The direct regression system as shown in Figure~\ref{fig:direct-pipeline} uses separate translation and rotation branches, global and cropped event representations, an event-spread scale cue, and physics-motivated synthetic corruptions. Both systems are evaluated by the same hidden-test protocol on real event-camera recordings and produce predictions for all test frames.

The comparison is intentionally system-level. The pipelines differ not only in pose formulation but also in event representation, network capacity, target-domain access, training objective, and temporal processing. Their performance difference therefore cannot be attributed causally to a single architectural choice. Instead, the comparison asks an operational question: under a common hidden-test protocol, how do two complete direct and geometry-based systems distribute their errors between translation and orientation?
We summarize the key contributions below:

\begin{itemize}
\item We document and compare two independently engineered event-based spacecraft pose-estimation systems under a common SPADES hidden-test evaluation, reporting translation, orientation, composite error, and frame coverage in consistent units.

\item We reveal a component-wise performance reversal concealed by the aggregate challenge score. The direct system reduces orientation error by 22.7\%, from 1.1767 to 0.9101~rad, whereas the modular system reduces relative translation error by 78.7\%, from 0.1335 to 0.0285.

\item Using archived configurations and an interpretable correspondence failure case, we identify explanations consistent with the observed error profiles and formulate a testable hybrid design: direct orientation estimation combined with geometry-constrained translation recovery. We present this combination as a hypothesis for future evaluation rather than as a validated method.

\end{itemize}

The 
{ implementation} and 
{model artifacts} will be made publicly available.
The remainder of this paper is organized as follows: Section~\ref{sec:related} reviews related work in event-based vision and pose estimation. Section~\ref{sec:spades} describes SPADES and the evaluation setting. Sections~\ref{sec:direct} and ~\ref{sec:modular} document the two submitted systems. Section~\ref{sec:results} reports the official results, configuration analysis and design implications, followed by concluding remarks in Section~\ref{sec:conclusion}.

%% file: sec/2_related.tex
\section{Related Work}
\label{sec:related}

\subsection{Event-based Spacecraft Pose Estimation}
Early work on uncooperative spacecraft pose estimation largely used frame-based imagery. SPEED~\cite{kisantal2020satellite} established the first large-scale synthetic benchmark, while SPEED+~\cite{park2022speed+} added hardware-in-the-loop imagery for evaluating the synthetic-to-real domain gap. Event cameras complement conventional cameras by asynchronously measuring brightness changes with low latency and high dynamic range~\cite{gallego2020event}. Jawaid et al.~\cite{10160531} demonstrated the relevance of event sensing for spacecraft imagery, and SPADES~\cite{rathinam2024spades} introduced paired synthetic and laboratory event sequences with pose supervision. Our study uses the SPARK challenge setting to examine how two complete systems transfer from simulated training data to physical event-camera recordings.

\subsection{Event-based Pose-Estimation Pipelines}
Transitioning from standard images to asynchronous event streams requires specialized processing architectures. Existing methods typically condense the event streams into spatial-temporal representations such as voxel grids~\cite{zhu2019unsupervised} or time surface maps~\cite{lagorce2017hots}. Once processed, the literature generally splits into two distinct pose estimation paradigms. A modular design predicts $2$D keypoints and uses their correspondences with known $3$D points in a PnP solver~\cite{chen2019satellite, chen2020end}. This structure exposes intermediate predictions and explicitly uses object geometry. A direct design instead regresses translation and rotation from learned features~\cite{proenca2020deep}, avoiding a discrete correspondence bottleneck but relying on the training distribution to learn metric pose cues. 

The two systems studied here instantiate these ideas differently. The modular system uses a $20$-bin event representation~\cite{zhu2019unsupervised}, HRNet keypoints, CycleGAN and DANN adaptation, PnP with RANSAC~\cite{fischler1981random}, and EKF smoothing~\cite{rauch1965maximum}. The direct system uses three-channel event accumulation, separate global and cropped views, two convolutional backbones~\cite{tan2021efficientnetv2, liu2022convnet}, a scale feature, and synthetic corruptions. We retain these end-to-end configurations because system performance emerges from precisely these interactions, the complete pipeline is the unit of comparison.

\begin{figure}[t]
  \centering
   \includegraphics[width=1\linewidth]{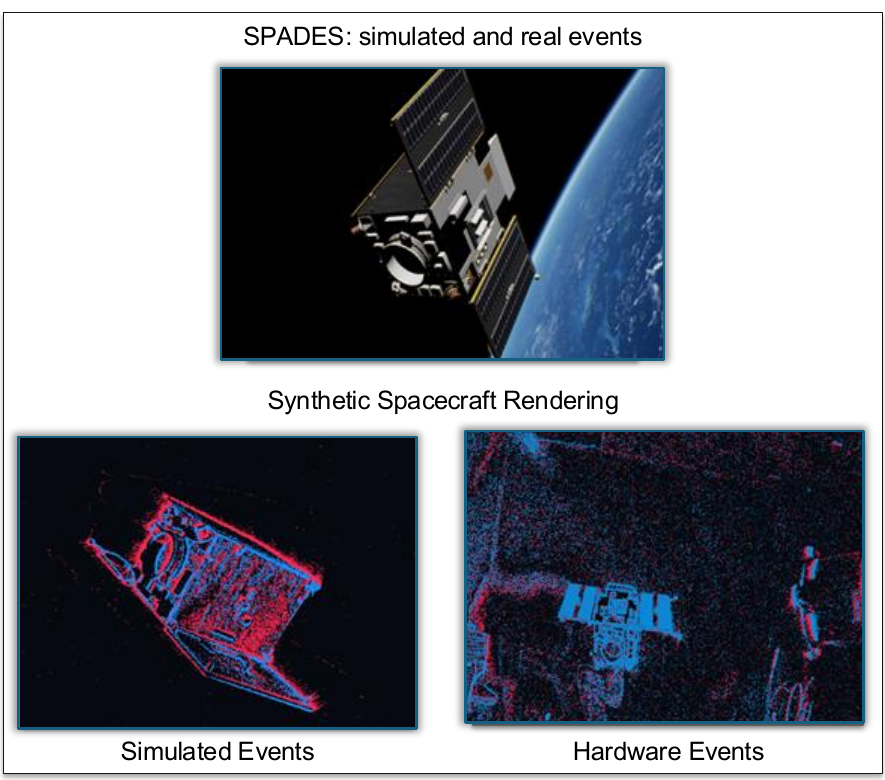}
  \caption{The synthetic-to-real setting of the SPADES benchmark, showing a
synthetic spacecraft rendering, simulated events, and events recorded with
physical hardware. The synthetic rendering on the top is reproduced
from SPADES~\cite{rathinam2024spades}.}
  \label{fig:SPADES}
\end{figure}

\subsection{Synthetic-to-Real Adaptation}
The synthetic-to-real gap may be addressed through input translation, adversarial feature alignment, test-time adaptation, augmentation, ensembling, or model-based rendering~\cite{park2024robust}. These techniques can be highly effective: the SPEC2021 competition analysis identifies adversarial training as a crucial ingredient in its winning SPEED+ systems~\cite{park2023spec}. Such pipelines are designed to maximize target-domain accuracy, and their results establish that stronger aggregate performance is attainable than with minimally adapted baselines.

That objective is distinct from the question addressed here. A leaderboard score does not reveal whether an improvement comes from the pose representation, the adaptation module, temporal post-processing, or selective confidence behavior. In this sense, adaptation and render-and-compare stages can act as black boxes for architectural diagnosis even when they are excellent predictors: they improve the final score without necessarily exposing which geometric cue succeeds or fails. Our modular system itself uses CycleGAN and DANN, but its explicit keypoints and PnP stage retain a geometric interface that can be inspected, the direct system instead tests corruption-based invariance without target pose labels.
These submissions occupy a deliberately different point from accuracy-maximizing ensembles and rendering pipelines: they are two instrumented systems that expose a question stronger leaderboard entries leave unresolved. Why does the ordering reverse between orientation and translation? The component-wise analysis turns that ranking into a testable design hypothesis combining the strongest measured component of each pipeline.

%% file: sec/3_spades.tex
\section{The SPADES Benchmark Suite}
\label{sec:spades}

Recent progress in spacecraft pose estimation has closely followed the development of increasingly realistic benchmark datasets. Each major dataset has addressed the limitations and shortcomings of the earlier benchmarks, particularly the synthetic-to-real domain gap. 
In this section, we review the key benchmarks (see Table~\ref{tab:datasets}) and explain why SPADES~\cite{rathinam2024spades} provides an appropriate setting for evaluating event-based spacecraft pose estimation.

\subsection{Spacecraft Pose Estimation Datasets}
\label{sec:dataset-evolution}
\begin{table}[t]
\centering
\caption{Evolution of spacecraft pose estimation benchmarks. SPADES is the only listed dataset combining event-based sensing, synthetic and real data, and full pose supervision.}
\label{tab:datasets}
\small
\setlength{\tabcolsep}{8pt}
\renewcommand{\arraystretch}{1.18}

\begin{tabular}{@{}lcccc@{}}
\toprule
\textbf{Dataset} & \textbf{Sensor} & \textbf{Synthetic} & \textbf{Real} & \textbf{Pose Labels} \\
\midrule

SPEED~\cite{kisantal2020satellite}
& Frame
& \checkmark
& \texttimes
& \checkmark \\

SPEED+~\cite{park2022speed+}
& Frame
& \checkmark
& \checkmark
& \checkmark \\

SEENIC~\cite{10160531}
& Event
& \checkmark
& \checkmark
& \texttimes \\

SPADES~\cite{rathinam2024spades}
& \textbf{Event}
& \checkmark
& \checkmark
& \checkmark \\

\bottomrule
\end{tabular}
\end{table}

The Spacecraft Pose Estimation Dataset (SPEED)~\cite{kisantal2020satellite} established an early large-scale benchmark using synthetic images of the Tango spacecraft from the Swedish PRISMA mission with ground-truth pose annotations. SPEED+ \cite{park2022speed+} later added real lightbox imagery, enabling direct study of the synthetic-to-real domain gap. However, both rely on conventional frame-based cameras.
Jawaid et al.~\cite{10160531} introduced SEENIC, which extended spacecraft pose estimation to event cameras and showed improved robustness to large lighting variations. Its main limitation is the lack of pose labels for both synthetic and real data, which limits quantitative evaluation of learned pose estimators.
SPADES~\cite{rathinam2024spades}, illustrated in
Figure~\ref{fig:SPADES}, contains simulated and hardware-recorded
event sequences of the physical ESA PROBA-$2$ mockup with full $6$-DoF pose annotations. Captured using a Prophesee Metavision EVK$4$ sensor~\cite{prophesee_evk4}, SPADES combines event-based sensing with labeled real-world data, making it well suited for studying synthetic-to-real generalization. It also serves as the basis of the SPARK 2026 Challenge.

\subsection{The SPADES Dataset}
SPADES~\cite{rathinam2024spades}, introduced by the University of Luxembourg, contains simulated and real event-based recordings of the PROBA-$2$ satellite. The synthetic split was generated in Unreal Engine by rendering a CAD model to RGB frames and converting them into asynchronous event streams using the International Centre for Neuromorphic Systems (ICNS) event simulator~\cite{joubert2021event}. The real split was captured at $1280\times720$ resolution using a Prophesee Metavision EVK$4$-HD sensor~\cite{prophesee_evk4} observing a physical PROBA-$2$ mockup measuring approximately $0.64 \times 0.24 \times 0.416\,\mathrm{m}$. Although the benchmark baseline accumulates raw events into three-channel pseudo-RGB frames, the underlying asynchronous streams $(x_i,y_i,t_i,p_i)$ can be discretized into other temporal representations, including the $3$-channel and $20$-bin representations used by our two systems (Sections~\ref{sec:direct} and~\ref{sec:modular}). 
The dataset contains 179,400 labeled synthetic poses and 16,930 real event samples with pose annotations maintained by the challenge evaluator. Challenge participants trained pose predictors using the released synthetic labels and submitted predictions for server-side scoring on the real sequences, the real test annotations were not available to the authors. 

In this work, SPADES is used to train and evaluate both systems. The modular submission additionally uses unlabeled target-domain observations for CycleGAN, DANN, and test-time batch-normalization adaptation, as described in Section~\ref{sec:modular}, the direct submission uses no real pose labels. The challenge returned aggregate hidden-test errors for the submitted configurations. Section~\ref{sec:results} analyzes these official results in common units and separates orientation from translation.

%% file: sec/4_direct.tex
\section{The Direct Regression System}
\label{sec:direct}
The direct submission maps an accumulated event representation to translation and rotation without explicit $2$D-$3$D correspondences. It uses a global view for translation and an adaptive crop for rotation, reflecting the design hypothesis that apparent scale and local structure provide different pose cues (see Figure~\ref{fig:direct-pipeline}).

\subsection{Event Windowing and Temporal Voxelization}
Following the event-windowing protocol described in SPADES \cite{rathinam2024spades}, we process the asynchronous event stream ${(x_i, y_i, t_i, p_i)}$ by accumulating events within a $400$-ms window centered around each label timestamp. This balances the motion informativeness and global spatial context against the accumulation blur. We adopt the $3$-channel (3C) pseudo-frame representation by splitting the temporal window into three equal sub-intervals mapped to red, green, and blue channels and accumulating events into a $320 \times 320$ spatial grid.
We extend this base representation with three preprocessing steps designed to reduce sensitivity to sensor noise. First, \emph{polarity-aware weighting} scales OFF events using the global ON/OFF ratio. Second, a \emph{linear time-decay weight} down-weights older events within the window. Third, a normalization sequence applies $2\times2$ max-pooling with stride 1, a $3\times3$ box blur and unsharp mask, log compression, a sigmoid ``squeeze,'' and per-channel percentile clipping to $[0,1]$.

\subsection{Dual-View Representation}
Motivated by the different cues used by translation and rotation~\cite{proenca2020deep}, we construct two voxel-grid representations from the same event window. This is a design choice rather than evidence that a shared encoder cannot model both quantities.
The global full-frame voxel $\mathbf{X}_g \in \mathbb{R}^{3\times320\times320}$ retains the full sensor field of view, preserving the global context and spacecraft's position and distance cues required for translation. The adaptive satellite-centered crop $\mathbf{X}_c \in \mathbb{R}^{3\times320\times320}$ is obtained by locating the median event coordinate as the satellite centroid and extracting a crop whose size is directly proportional to the spatial spread of the event cloud, isolating the fine structural detail required for rotational estimation while also remaining scale-normalized.

The translation branch also receives a scalar apparent-scale feature $s_c$, computed from the event-cloud spread:
\begin{equation}
\sigma_{\text{crop}} = \tfrac{1}{2}(\sigma_x + \sigma_y), \qquad s_c = \frac{\sigma_{\text{crop}}}{1280},
\label{eq:scalehint}
\end{equation}
where $\sigma_x$ and $\sigma_y$ are the standard deviations of active event coordinates and 1280 is the native horizontal resolution. Under fixed object extent and visibility, image scale is related to inverse depth through the pinhole model. Event-cloud spread is only a proxy: it can also vary with motion, orientation, contrast thresholds, and background activity. We therefore describe $s_c$ as an auxiliary cue rather than a solution to monocular depth ambiguity.

\subsection{Network Architecture}
The direct regression pipeline consists of two independent CNN backbones with no shared weights, allowing each branch to specialize to the structure of its respective input. The translational branch uses an EfficientNet-V2-S backbone~\cite{tan2021efficientnetv2} that is pretrained on ImageNet~\cite{deng2009imagenet}, applied to $\mathbf{X}_g$. After global average pooling, the result is a 1280-dimensional feature vector that is concatenated with the scalar scale hint $s_c$ before being passed through a two-layer multilayer perceptron with rectified linear unit (ReLU) and dropout to regress translation $\hat{t} \in \mathbb{R}^3$.

The rotational branch utilizes a ConvNeXt-Tiny backbone~\cite{liu2022convnet}, also pretrained on ImageNet~\cite{deng2009imagenet}, applied to the cropped voxel $\mathbf{X}_c$. The resulting $768$-dimensional pooled vector is then passed through a separate two-layer multilayer perceptron to produce a raw quaternion $\tilde{q} \in \mathbb{R}^4$ which is then normalized to unit length ($\hat{q} = \tilde{q}/\|\tilde{q}\|$) to ensure all predictions correspond to valid rotations on $SO(3)$.

At inference, the submitted configuration evaluates five input scales and combines the resulting rotation estimates using a geodesic mean~\cite{markley2007averaging}, as summarized in Figure~\ref{fig:direct-pipeline}.

\begin{figure*}[!t]
  \centering
  \includegraphics[width=0.95\textwidth]{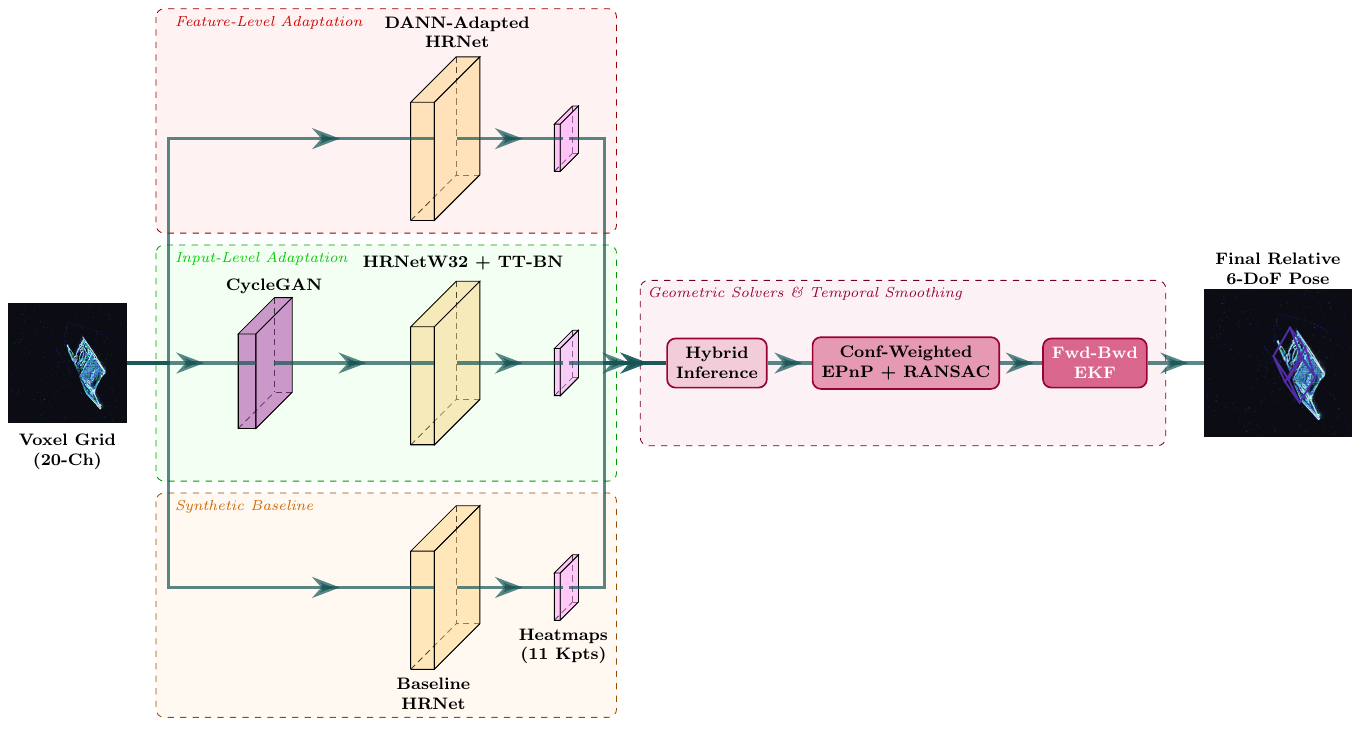}
  \caption{Modular keypoint-tracking architecture. CycleGAN+TT-BN and DANN provide the two adapted keypoint branches used by the deployed hybrid, the unadapted HRNet is shown as a reference configuration. Confidence-weighted EPnP with RANSAC is solved separately for each adapted branch before component selection and EKF smoothing.}
  \label{fig:modular-pipeline}
\end{figure*}

\subsection{Training Objective}
We supervise the pipeline using a weighted variant of the ESA composite pose score, $\mathcal{L}_{\text{pose}} = E_T + \lambda_R E_R$, where $E_T$ is relative translation and $E_R$ is geodesic rotation error. We set $\lambda_R = 2.0$, selected from $\{1.0,1.5,2.0,3.0\}$ on the synthetic validation split. This choice intentionally emphasizes rotation and must be considered when interpreting the component-wise test results. The direct system is trained for 14 epochs on a single NVIDIA A30 GPU using AdamW~\cite{loshchilov2019decoupled}, mixed-precision computation (FP16), and a batch size of 16. The learning rate is set to $5\times10^{-5}$ for epochs 1--7 and reduced to $1\times10^{-5}$ for epochs 8--14. We select the checkpoint from epoch 8 using the lowest ESA score on a synthetic validation split constructed by holding out approximately 20\% of complete trajectories, thereby preventing within-trajectory leakage. Both backbones are initialized with ImageNet-pretrained weights, while all regression layers are trained from scratch.

\subsection{Physically Motivated Augmentation}
\label{sec:augmentation}

The direct submission uses synthetic corruptions intended to approximate classes of artifacts encountered in the SPADES laboratory setting rather than adversarial feature alignment. Table~\ref{tab:augs} lists the twelve transforms applied independently to $\mathbf{X}_g$ and $\mathbf{X}_c$. Because these transforms were selected with knowledge of the target acquisition setting, they constitute a task-specific prior even though no real pose labels are used.
\begin{table}[t]
\centering
\caption{Physics-grounded augmentation suite applied to $\mathbf{X}_g$ and $\mathbf{X}_c$ during training.}
\label{tab:augs}
\small
\begin{tabular}{@{}llc@{}}
\toprule
\textbf{Name} & \textbf{Simulated artifact} & \textbf{$p$} \\
\midrule
Blue-floor bias & Lab floor/fixture reflections & 0.7 \\
Edge enhancement & High-activity region sharpening & 1.0 \\
Organic noise & Structured sensor/readout noise & 0.7 \\
Vignette & Corner darkening / gain falloff & 0.5 \\
Chromatic aberration & Channel misalignment / warping & 0.5 \\
Optical smear & Anisotropic feature bloom & 0.5 \\
Debris streaks & Transient streaks in FOV & 0.3 \\
Background bleed & Low-signal background activation & 0.5 \\
Secondary lights & Extra light sources / reflections & 0.3 \\
Lens flare & Radial/ghost flare artifacts & 0.3 \\
Brightness jitter & Global exposure/contrast change & 0.7 \\
JPEG compression & Codec artifacts and banding & 0.7 \\
\bottomrule
\end{tabular}
\end{table}
In addition to these photometric transforms, we apply in-plane rotational jitter: with probability 0.5, we sample an angle $\theta \in [-45^\circ, 45^\circ]$ and rotate both $\mathbf{X}_g$ and $\mathbf{X}_c$, composing the ground-truth quaternion with the corresponding camera-frame rotation to preserve label consistency.

\subsection{Domain Anchor Regularization}
\label{sec:anchor}

To encourage corruption-stable features without real pose labels, we apply a domain-anchor regularizer to both branches. Let $\phi_g(\mathbf{X}_g)\in\mathbb{R}^{1280}$ and $\phi_c(\mathbf{X}_c)\in\mathbb{R}^{768}$ denote the respective penultimate pooled features. At each training step, we evaluate each clean view and a heavily corrupted counterpart generated with blue-floor bias, organic noise, and low-quality JPEG compression. The branch-normalized penalty is
\begin{equation}
\mathcal{L}_{\text{anchor}} =
\sum_{b\in\{g,c\}}\frac{1}{d_b}
\left\|\phi_b(\mathbf{X}_b)-\phi_b(\widetilde{\mathbf{X}}_b)\right\|_2^2,
\label{eq:anchor}
\end{equation}
where $d_g=1280$ and $d_c=768$. We add it to the objective as $\mathcal{L}_{\text{total}}=\mathcal{L}_{\text{pose}}+\lambda_{\text{anchor}}\mathcal{L}_{\text{anchor}}$, with $\lambda_{\text{anchor}}=0.1$. This encourages each backbone to produce similar features under the selected synthetic corruptions, it does not by itself establish invariance to the full real domain.

%% file: sec/5_modular.tex
\section{The Modular Keypoint-Tracking System}
\label{sec:modular}

The modular submission decomposes $6$-DoF pose estimation into detecting $2$D spacecraft keypoints, establishing correspondences with known $3$D locations, and recovering pose with a geometric solver. Its submitted configuration combines HRNetW$32$ keypoints, CycleGAN~\cite{zhu2017unpaired} and DANN~\cite{ganin2016domain} adaptation, confidence-weighted EPnP with RANSAC, and forward-backward EKF smoothing (see Figure~\ref{fig:modular-pipeline}).

\subsection{Event Representation}
For each $400$-ms label-centered SPADES interval, raw events are rasterized at the native $H\times W=720\times1280$ sensor resolution and discretized into $B=20$ temporal bins. Each bin accumulates polarity-weighted event counts, producing a $20\times720\times1280$ voxel grid~\cite{zhu2019unsupervised} before the standard spatial resizing used by HRNet. This differs from the $3$-channel, $320\times320$ pseudo-frame used in Section~\ref{sec:direct}: the finer temporal discretization preserves motion cues used by the keypoint detector to resolve structural edges and corners from sparse events

\subsection{Keypoint Detection and Domain Adaptation}
An HRNetW32 backbone~\cite{8953615,wang2020deep}, initialized with ImageNet weights~\cite{deng2009imagenet} and fine-tuned on the SPADES \cite{rathinam2024spades} synthetic data, predicts $2$D spatial heatmaps and confidence scores for 11 spacecraft keypoints on the PROBA-$2$ satellite. HRNet's parallel high-to-low resolution branches preserve the spatial precision required for accurate heatmap localization of a small, geometrically constrained object in a sparse event field.
To reduce the synthetic-to-real gap, the modular system accesses unlabeled target-domain observations at two levels. A CycleGAN~\cite{zhu2017unpaired} translates synthetic voxel frames toward the target appearance without paired samples, and test-time batch-normalization adaptation~\cite{su2024unraveling} updates normalization statistics on target inputs. At the feature level, DANN~\cite{ganin2016domain} uses a gradient-reversal domain classifier to align synthetic and target features. Thus, the system uses no real pose labels but is not target-data-free, this distinction is central to interpreting its comparison with the direct submission.

\subsection{Pose Recovery, Hybrid Inference, and Temporal Smoothing}
Given $2$D keypoints and confidence scores, $6$-DoF pose is recovered using confidence-weighted EPnP~\cite{item_34721c4a90c2421c9df615a96127a4a2} with RANSAC~\cite{fischler1981random}. This stage ties the output to known spacecraft geometry and exposes interpretable intermediate predictions. It can nevertheless fail when too few correspondences remain accurate, as illustrated by the case study in Section~\ref{sec:results}.
The submitted modular system applies confidence-weighted EPnP with RANSAC independently to the $2$D keypoints predicted by the CycleGAN+TT-BN and DANN-adapted branches. The unadapted synthetic HRNet shown in Figure~\ref{fig:modular-pipeline} is a reference configuration and does not contribute to the deployed hybrid. The final per-frame pose retains the orientation estimate $\hat{\mathbf q}$ from the CycleGAN+TT-BN solver and the translation estimate $\hat{\mathbf t}$ from the DANN solver, after which the pose sequence is refined offline with a forward-backward EKF~\cite{rauch1965maximum}. This sequence-level temporal context differs from the frame-independent direct system and is another reason the final scores should be interpreted as a comparison of complete deployed pipelines rather than isolated regressors and geometric solvers.

\subsection{Implementation Details}
All models are trained on an institutional computing cluster using a single NVIDIA A100 80GB GPU. HRNetW32 is fine-tuned on the SPADES synthetic split with a combined heatmap regression and confidence loss. CycleGAN is trained for 200 epochs using Adam ($\beta_1 = 0.5$, $\beta_2 = 0.999$, lr $= 2 \times 10^{-4}$, linearly decayed after epoch 100). DANN uses SGD (lr $= 1 \times 10^{-3}$, gradient reversal $\lambda = 1.0$). All voxel grids are normalized to $[-1, 1]$ before input.

%% file: sec/6_experiments.tex
\section{Experiments and Results}
\label{sec:results}
Pose-supervised training uses the labeled SPADES~\cite{rathinam2024spades} synthetic split. The modular system also accesses unlabeled target-domain samples through CycleGAN, DANN, and test-time normalization, the direct system uses target-informed synthetic corruptions but no real pose labels. For frame $i$, SPADES and SPARK use
\begin{equation}
e_T^i=\frac{\|\hat{\mathbf t}_i-\mathbf t_i\|_2}{\|\mathbf t_i\|_2},\qquad
e_R^i=2\arccos\!\left(\left|\langle\hat{\mathbf q}_i,\mathbf q_i\rangle\right|\right),
\label{eq:metrics}
\end{equation}
where $e_T^i$ is a unitless relative translation error and $e_R^i$ is the quaternion geodesic error in radians. The challenge score averages their per-frame sum over the evaluated frames, $E_P=N^{-1}\sum_i(e_T^i+e_R^i)=E_T+E_R$.

The SPADES paper displays $100E_T$ as a percentage and $180E_R/\pi$ in degrees, but computes $E_P$ after converting to the native fraction and radian units. We use the native units throughout. For example, its real direct baseline reports $E_T=5.13\%$ and $E_R=81.13^\circ$, these become $0.0513$ and $1.4160$ rad, giving $E_P=1.4673$. Thus, the SPADES and challenge scores use the same metric after unit normalization.
The challenge records provide official aggregate scores for the submitted configurations, with real annotations held by the evaluator. We report exact observed differences and analyze the complete systems as deployed. Their differences in representation, capacity, adaptation, and temporal inference are not nuisances to erase, they define the two engineering strategies under comparison.
\subsection{The Results}
\begin{table}[t]
\centering
\caption{SPADES real-data results in common units: relative translation as a fraction and orientation in radians. Coverage is the percentage of frames scored. Bold marks the lower error among the two SPARK submissions.}
\label{tab:results}
\small
\resizebox{\columnwidth}{!}{%
\begin{tabular}{@{}llcccc@{}}
\toprule
\textbf{System} & \textbf{Configuration} & \textbf{Cov.} & $E_R\downarrow$ & $E_T\downarrow$ & $E_P\downarrow$ \\
\midrule
\multicolumn{6}{l}{\textit{Published SPADES baselines~\cite{rathinam2024spades}}} \\
Direct  & MobileNet-V3-L~\cite{howard2019searching}, no adapt.     & 73.32\% & 1.4160 & 0.0513 & 1.4673 \\
Modular & HigherHRNet~\cite{cheng2020higherhrnet} + BPnP~\cite{chen2020end}, no adapt. & 17.27\% & 1.3785 & 0.0334 & 1.4119 \\
\midrule
\multicolumn{6}{l}{\textit{Independently developed SPARK submissions}} \\
Modular & HRNet, no adaptation & All & 1.9137 & 0.0412 & 1.9549 \\
Modular & Submitted full system & All & 1.1767 & \textbf{0.0285} & 1.2052 \\
Direct  & Submitted full system & All & \textbf{0.9101} & 0.1335 & \textbf{1.0436} \\
\bottomrule
\end{tabular}%
}
\end{table}

\subsection{Archived Configuration Comparisons}

\begin{table}[t]
\centering
\caption{Official composite scores for archived challenge configurations. Each row records a submitted system configuration.}
\label{tab:ablation}
\resizebox{\linewidth}{!}{%
\begin{tabular}{@{}lllc@{}}
\toprule
\textbf{System} & \textbf{Adaptation} & \textbf{Augments} & $E_P\downarrow$ \\
\midrule
Modular & CycleGAN + DANN & None & 1.3347 \\
\midrule
Direct & None & None & 1.7299 \\
Direct & Domain anchor & None & 1.4664 \\
Direct & None & Illum./noise & 1.2879 \\
Direct & None & Spatial/artifact & 1.2293 \\
Direct & Domain anchor & Full suite & \textbf{1.0436} \\
\bottomrule
\end{tabular}%
}
\end{table}

Table~\ref{tab:results} shows complementary rather than uniformly ordered performance. Relative to the submitted modular system, the direct system reduces orientation error by 22.7\% ($1.1767\rightarrow0.9101$ rad). Conversely, the modular system reduces relative translation error by 78.7\% ($0.1335\rightarrow0.0285$), or by a factor of 4.7. The direct system's composite score is 13.4\% lower ($1.2052\rightarrow1.0436$), not 22.7\% lower.
This distinction matters because $E_P$ adds quantities with different units and numerical scales. Rotation dominates these submissions' aggregate score, so the composite ranking favors the direct system despite its substantially larger translation error. On this benchmark, the observed profiles suggest the direct submission for attitude-weighted use and the modular submission for position-weighted use, they do not establish that either architectural family is generally preferable.

The translation gap is consistent with two design-level hypotheses. First, the modular system constrains translation through known $3$D correspondences, whereas background activity can inflate the direct system's event-spread cue. Second, the direct objective uses $\lambda_R=2$, emphasizing orientation. The available aggregate records cannot isolate either mechanism, so we treat them as explanations to test rather than causal conclusions.
The published SPADES rows provide historical reference points, but they use different backbones, optimization procedures, and evaluation coverage. The SPADES direct and modular baselines return confident poses for only 73.32\% and 17.27\% of the real frames, respectively, the paper explicitly notes that confidence filtering removes the remaining samples. Their errors therefore describe different, selectively retained subsets and cannot be compared directly with all-frame challenge submissions. Likewise, the unadapted HRNet row is only a starting configuration for the submitted modular system, not a general baseline for all keypoint methods.

The archived direct configurations progress from 1.7299 without adaptation to 1.4664 with the domain anchor, 1.2879 with illumination/noise transforms, 1.2293 with spatial/artifact transforms, and 1.0436 with the anchor and full suite. The first group contains brightness, blue-floor, vignette, secondary-light, flare, and organic-noise effects, the second contains edge, chromatic, smear, debris, bleed, and JPEG effects. These non-nested submissions support the complete recipe without isolating individual transforms.

Because the historical challenge log retained only the aggregate leaderboard score $E_P$ for these intermediate submissions, Table~\ref{tab:ablation} reports composite progression. Component-level decomposition is therefore restricted to the primary configurations in Table~\ref{tab:results}.
The modular score of 1.3347 records the intermediate CycleGAN+DANN configuration. The final score of 1.2052 in Table~\ref{tab:results} includes the submitted inference and post-processing pipeline. The 0.1295 difference is associated with all intervening changes and does not isolate the effect of hybrid component selection or EKF smoothing.

\subsection{Qualitative Case Study}
\begin{figure}[t]
  \centering
   \includegraphics[width=1\linewidth]{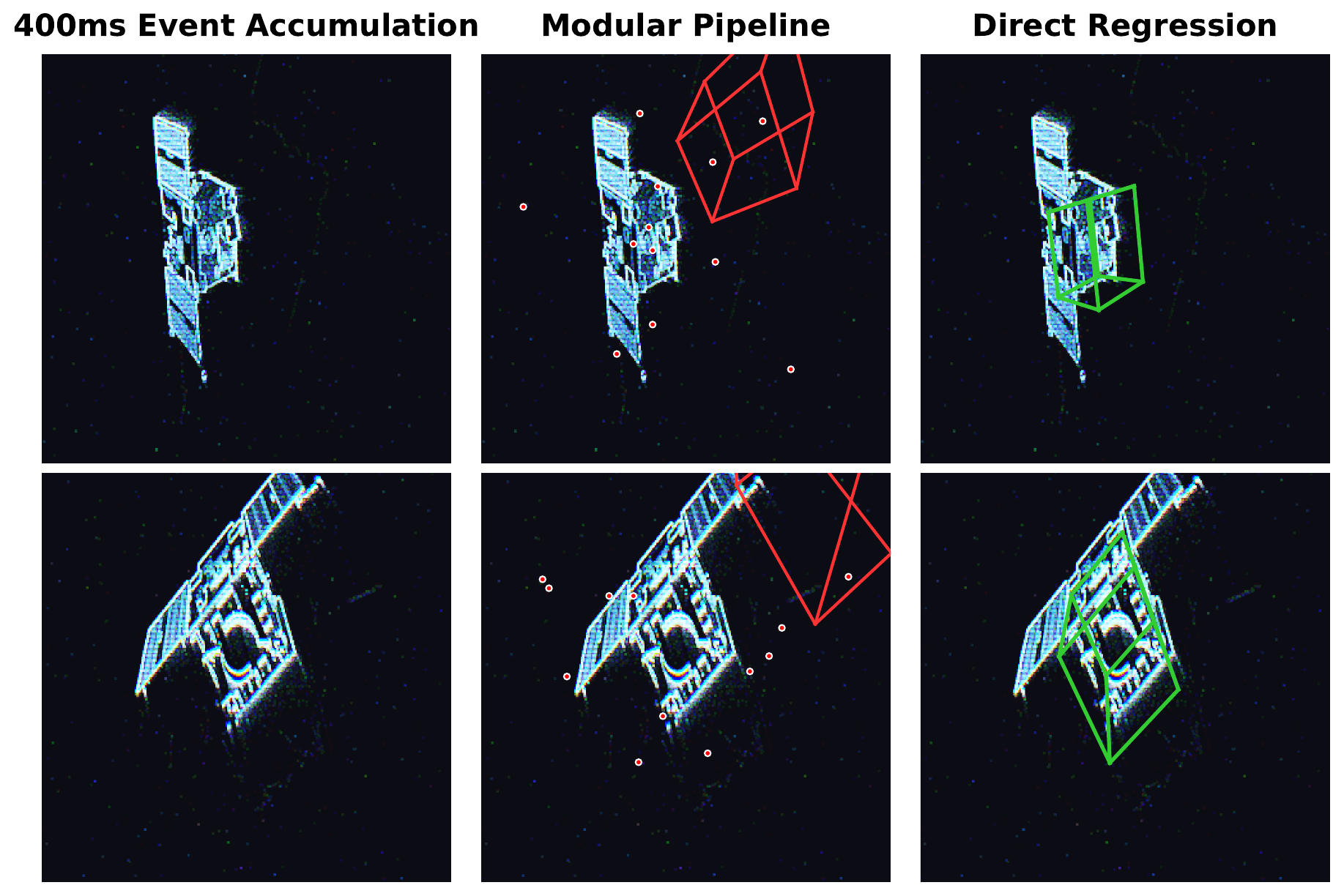}
  \caption{Two representative frames from a challenging real SPADES sequence. Red markers denote modular HRNet keypoints, the red wireframe is the resulting modular EPnP pose projection, and the green wireframe is the direct-regression pose projection. Background activity produces inconsistent correspondences and a failed modular pose, while direct regression remains aligned with the spacecraft.}
  \label{fig:qualitative}
\end{figure}

As in Figure~\ref{fig:qualitative}, the two selected frames, background activity coincides with inconsistent HRNet detections, too few valid correspondences, and a displaced modular pose, the direct projection remains better aligned. This illustrates a plausible failure mode but neither measures its frequency nor explains the aggregate gap.

\subsection{Threats to Validity and Study Scope}

The unit of analysis is the complete deployed pipeline rather than any individual component. The two systems differ in their event representations, neural backbones, training objectives, domain-adaptation strategies, target-domain access, geometric constraints, and temporal processing. Consequently, the observed translation--orientation reversal cannot be attributed causally to direct regression, keypoint-based recovery, or any other isolated design choice. In particular, the modular system uses unlabeled target-domain observations and offline temporal smoothing, whereas the direct system relies on target-informed synthetic corruptions and frame-independent inference. We therefore interpret the results as a comparison of two operational system designs, not as a controlled ablation of architectural families.
The hidden-test evaluator provides aggregate translation, orientation, and composite errors, but the real test labels and per-frame error records are unavailable. We therefore cannot report confidence intervals, statistical significance, per-trajectory variability, or performance as a function of range, motion, illumination, or event density. Moreover, the composite score adds a unitless relative translation error to a rotation error measured in radians. Because these components have different numerical scales, the aggregate ranking emphasizes orientation for the evaluated submissions. We consequently report the component errors and frame coverage alongside the composite score.

%% file: sec/7_conclusion.tex
\section{Conclusion}
\label{sec:conclusion}


We present two independently developed systems for synthetic-to-real, event-based spacecraft pose estimation under the official SPARK $2026$ hidden-test protocol. The direct-regression system reduced orientation error by $22.7$\% and achieved a $13.4$\% lower composite score, whereas the modular keypoint-and-geometry system reduced relative translation error by $78.7$\%. This component-wise reversal demonstrates that an aggregate pose score can conceal application-relevant differences between translation and orientation performance.
Because the pipelines differ in representation, adaptation, architecture, objective, and temporal processing, the observed reversal cannot be attributed to any single design choice. Nevertheless, their complementary error profiles motivate a concrete hypothesis for future evaluation: combining direct orientation estimation with geometry-constrained translation recovery and joint pose refinement. Future work should implement this hybrid and evaluate its components through controlled ablations across additional spacecraft and sensing conditions. More broadly, pose-estimation benchmarks should report translation error, orientation error, and frame coverage alongside any composite score.